\documentclass[runningheads]{llncs}

\usepackage{graphicx}
\usepackage{tabularx}

\usepackage{soul}
\usepackage{hyperref}
\usepackage{pgfplots}
\usepackage[utf8]{inputenc}
\usepackage[T5]{fontenc}
\usepackage{float}
\usepackage{tikz}
\usepackage{longtable}
\usepackage{supertabular}

\usetikzlibrary{patterns}

\usepackage{epstopdf}
\usepackage[utf8]{inputenc}

\usepackage{hyperref}
\usepackage{xstring}

\usepackage{color}

\begin{document}
\title{Detecting Spam Reviews on Vietnamese E-commerce Websites}

\author{
Co Van Dinh\inst{1,2} \and
Son T. Luu\inst{1,2} \and
Anh Gia-Tuan Nguyen\inst{1,2}}

\institute{University of Information Technology, Ho Chi Minh City, Vietnam \and
Vietnam National University Ho Chi Minh City, Vietnam \\
\email{19521293@gm.uit.edu.vn, sonlt@uit.edu.vn, anhngt@uit.edu.vn}}

\titlerunning{ }
\authorrunning{ } 
\maketitle              

\begin{abstract}
The reviews of customers play an essential role in online shopping. People often refer to reviews or comments of previous customers to decide whether to buy a new product. Catching up with this behavior, some people create untruths and illegitimate reviews to hoax customers about the fake quality of products. These are called spam reviews, confusing consumers on online shopping platforms and negatively affecting online shopping behaviors. We propose the dataset called ViSpamReviews, which has a strict annotation procedure for detecting spam reviews on e-commerce platforms. Our dataset consists of two tasks: the binary classification task for detecting whether a review is spam or not and the multi-class classification task for identifying the type of spam. The PhoBERT obtained the highest results on both tasks, 86.89\%, and 72.17\%, respectively, by macro average F1 score. 

\keywords{spam reviews, text classification, deep neural models, transformer models, dataset, annotation guidelines}

\end{abstract}

\section{Introduction}
\label{intro}
Vietnam has witnessed strong growth in e-commerce in recent years. Many Vietnamese online trading platforms are constructed and attract consumers. Online shopping is now popular in people's daily routines because of its convenience and flexibility. However, besides the advantages of online shopping, the rise of fake products and fraud qualifications in online trading concerns customers and shop owners. 

Customer reviews play an essential part in the behaviors of consumers in online shopping. Customer reviews express their opinions, emotions, and attitudes about products, and these opinions affect other customers in deciding whether to buy a product or not. If a customer wants to buy a product in an online shop, they tend to refer to reviews of previous customers about that product. Catching up with this behavior, some people create spam reviews, which are illegitimate means and untruth facts about the actual quality of products to confuse the consumers to boost the financial business or fame of an individual or organization \cite{jindal2007review}. Therefore, detecting these spam reviews will protect both sellers and customers from the risk of low-quality products and preserve the reputation of the sellers. 

Our purpose in this paper is to propose a method to detect spam reviews about products on online shopping platforms. First, we constructed a corpus for spam detection from users' reviews by texts. Second, we use machine learning approaches to build classification models for detecting spam comments and evaluate classification models' performances on the constructed dataset.

The paper is structured as follows. Section \ref{intro} introduces our works. Section \ref{related_works} takes a survey about relevant research on the problem of online spam detection. Section \ref{dataset} describes the data creation process and gives an overview of our dataset. Section \ref{methodologies} introduces our approaches for the online spam reviews detection problem by machine learning and deep learning. Section \ref{results} displays our empirical results and analysis of the performances of classification models. Finally, Section \ref{conclusion} concludes our research and proposes future works.
\section{Related Works}
\label{related_works}
Preliminary research about spam reviews showed how to construct a classification model for detecting whether a user's reviews are spam or not and the difficulty in detecting spam reviews from both the content of the review and the reviewers \cite{jindal2007review}. The challenge of identifying spam reviews comes from reviewers' behavior when they try to create spam content just like other innocent reviewers. Therefore, \cite{jindal2008opinion} proposes three aspects of spam reviews to claim the problem of spam reviews. 

Many approaches are applied for the task of spam reviews detection, including patterns and rules \cite{xie2012review,li2017bimodal,li2015analyzing}, machine learning and deep learning approaches \cite{ren2016deceptive,8350457}, linguistic features \cite{li-etal-2014-towards,ott-etal-2011-finding}. Overall, the dataset is the key point for training and evaluating the classification models applied in the opinion spam detection task. Available datasets for detecting spam reviews are introduced by \cite{8678638}. 

In Vietnamese, there are several datasets about user reviews on e-commerce platforms, such as the dataset about phone and restaurant reviews \cite{nguyen2018vlsp,10.1145/3446678}, the smartphone feedback datasets \cite{10.1007/978-3-030-82147-0_53,nguyen2021span}, and the complaining detection on e-commerce websites dataset \cite{nguyen2021vietnamese}. However, there is no particular dataset for spam review detection on Vietnamese e-commerce websites yet. Hence, our motivation is to construct a dataset for detecting spam reviews on Vietnamese E-commerce platforms.
\section{The Dataset}
\label{dataset}
\subsection{Dataset Creation Process}

We collected data from leading online shopping platforms in Vietnam. Then, we select some of the most recent selling products for each product category and collect up to 15 reviews per product. After collecting, we get a dataset of 19,868 product reviews which contains the number of star reviews, comments about the product, and the link to that product. Subsequently, we construct the annotate guideline and annotate for the corpus. 

Our data annotating process is impressed from the MATTER framework \cite{pustejovsky2012natural}. The annotation process consists of two phases. The first is the training phase for the annotators. The annotators read a guideline describing the meaning of labels, a sample review, and some examples of specific cases. The annotators will read the guideline and annotate 300 random samples in the dataset. Then, we calculate and evaluate the average inter-annotator agreement. If the inter-annotator agreement is satisfied, we move to the second phase, which is the annotation phase. In contrast, we re-train the annotators and update the annotation guidelines. According to \cite{di-eugenio-2000-usage}, the inter-annotator agreement calculated by Cohen's Kappa index \cite{cohen1960coefficient} is acceptable when higher than 0.5.

In the second phase, the annotators will be provided with a complete dataset and annotate on this dataset. There are all three annotators during the annotation phases, and the final label of the dataset will be decided by voting for the most assigned label. To ensure objectivity when annotating the dataset, we keep annotators annotating independently. 

\subsection{Annotation Guidelines}
Our dataset comprises two tasks. The first task determines whether the reviews are spam or not spam (\textbf{Task 1}), and the second task indicates the types of spam reviews (\textbf{Task 2}). The dataset contains two labels: SPAM and NO-SPAM. For each spam review, we label one of three types of spam labels \cite{jindal2008opinion}. The label of a review is described as follows:

\textbf{NO-SPAM}: Reviews labeled with this label are regular reviews, true to the product's reality. Reviews like these provide helpful information for buyers to get an overview of the product before deciding whether to buy it or not.

\textbf{SPAM}: Reviews labeled with this label are reviews that are entirely or partially untrue about products sold on e-commerce sites. Reviews like these often make it easier to sell products or hurt the sales and reputation of stores and provide inaccurate or unhelpful information. According to \cite{jindal2008opinion}, we divided the labels for the reviews as spam into three labels:
\begin{itemize}
    \item \textbf{SPAM-1 (fake review)}: These reviews mislead customers by giving negative review comments to the product to damage the reputation of the store selling the product or giving an excellent review to the product in order to attract customers for the product and the shop even though the product is not relevant.
    
    \item \textbf{SPAM-2 (review on brand only)}: These reviews do not comment specifically on the product but only on the brand, manufacturer, or seller of the product. Although these reviews can be informative for product buyers, they are often negative and considered spam.
    
    \item \textbf{SPAM-3 (non-review)}: These are reviews whose commentary is not about the product or anything related to the product. These reviews tend to promote another product, get commissions from an e-commerce site, or have no purpose.
\end{itemize}

\begin{table}[H]
    \centering
    \caption{Several example reviews and instruction for annotating labels}
    
    \resizebox{1\textwidth}{!}{
    \begin{tabular}{p{8cm}|c|c|p{4cm}}
        \textbf{Comments} & \textbf{Type} & \textbf{Stars} & \textbf{Explanation} \\
        \hline
        Gia Vị Lẩu Haidilao (Hải Đế Lao) Chắc hẳn ai đi Trung Quốc cũng 1 lần ghé ăn nhà hàng lẩu Haidelao siêu siêu ngon với vị lẩu tuyệt hảo. Độ ngon độ hot của thương hiệu này thì không phải bàn cãi nữa, chỉ cần 1 gói gia vị lẩu chế được 1-2 nồi nước ngon ngọt siêu thơm. Sản phẩm gia vị độc đáo dùng. (\textit{\textbf{English}: Haidilao Hot Pot Seasoning - Surely everyone who goes to China will visit once at the super delicious Haidelao hotpot restaurant with excellent hotpot taste. The deliciousness and hotness of this brand is not in dispute anymore, just 1 pack of hot pot spices can make 1-2 pots of super delicious. This is a unique seasoning product.}) & SPAM-1 & 5 & This comment was created in order to advertise for the brand named Haidilao. It does not mention to any aspect of the product. \\
        \hline
        JBL thương hiệu có tiếng về loa. Cảm ơn Tiki đã giao hàng thành công, mặc dù chậm hơn 1 ngày so với lịch hẹn nhưng mình hiểu vì lý do thời tiết trời mưa. Chúc Tiki thành công và phát triển mạnh. (\textit{\textbf{English}: JBL brand is famous for speakers. Thank you Tiki for successful delivery, although 1 day later than scheduled, I understand because of the rainy weather. Wish Tiki success and thrive.}) & SPAM-2 & 5 & This review only focuses on the specific brand, which does not give any evaluation of the product. \\
        \hline
        ây đẹp lắm shopppppppppppppppppppp. Chất đẹp mịn sang. Nhận hàng ưng cực kì. Giá cả hợp lýyyyyyyyyyyyyyyyyyyyyyyyyyyyyyyy. Sẽ ủng hộ shopppppp dài dài ạ (\textit{\textbf{English}: very beatiful, shoppppppppppppppppppp. Quality fabric is beautiful, smooth, luxurious. Received very satisfied. Reasonable priceeeeeeeeeeeeeeeeeeeee. Will support shopppppp for a long time}) & SPAM-3 & 5 & This comment is not relevant to any aspect about the product. In this comment, the users use the word with duplicated characters to improve the comment length. However, this is not used to give any valuable opinion about the product quality. \\
        \hline
        Phấn mịn, bám da tốt. Mình sử dụng màu Peach hơi trắng hơn da mình. Không biết mấy màu khác thì sao. (\textit{\textbf{English}: The powder is smooth and adheres well to the skin. I use Peach color which is slightly whiter than my skin. I don't know about other colors}) & NO SPAM & 5 & This comment express the opinion of the users about the product quality. \\
        \hline
        Sản phẩm rất tệ, có mùi mật ong nhưng nhỏ trực tiếp lên đường đi của kiến để cả ngày chẳng thấy con kiến nào bu vào. Mà chúng còn làm thêm đường khác để đi. Phải rải thêm 1 ít sữa bột thì kiến mới bu vào. Nhưng cuối cùng kiến vẫn đi khắp nhà. Mình sử dụng dũng mấy ngày kquả rất bực nên mới đánh gía (\textit{\textbf{English}: The product is very bad, has a smell of honey but drops directly onto the ant's path so that no ants can come in all day. But ants also make another way to go. Have to sprinkle a little more milk powder to get the ants in. But in the end ants still go around the house. I used it for a few days and the results were so frustrating that I gave a review}) & NO SPAM & 1 & This comment gives the opinion of the consumer about the specific product. It can be seen that the attitude of the consumer with the product is matched with the rating stars he/she gave. Hence, this is a good review of the product. \\
        \hline
    \end{tabular}
    }
    
    \label{tbl_example}

\end{table}

In addition, Table \ref{tbl_example} describes several examples of reviews from users and the explanation for choosing the label for each review. For each review, annotators choose one suitable label.

\subsection{Inter-annotators Agreement and Discussion}
We have three different annotators to annotate the dataset. We let those three annotators work independently on the sample to measure the inter-annotator agreement in the training phase. Then, we calculate the inter-annotator agreement in pairs of annotators by the Cohen's Kappa index \cite{cohen1960coefficient}.

\begin{figure}[H]
    \centering
    \begin{minipage}[t]{.44\textwidth}
        \begin{table}[H]
        \caption{Inter-annotator agreement of three annotators A1, A2 and A3 on the two tasks. Annotators are working independently}
        \centering
        \begin{tabular}{|c|c|c|c|c|c|c|c|}
            \hline
            & \multicolumn{3}{|c|}{Task 1} & \multicolumn{3}{|c|}{Task 2}\\
            \hline
            & A1 & A2 & A3 & A1 & A2 & A3 \\
            \hline
            A1 & - & 0.36 & 0.34 & - & 0.41 & 0.37\\
            \hline
            A2 & - & - & 0.61 & - & - & 0.57 \\
            \hline
            A3 & - & - & - & - & - & - \\
            \hline
            \textit{Average $\kappa$} & \multicolumn{3}{|c|}{0.43} & \multicolumn{3}{|c|}{0.44} \\
            \hline
        \end{tabular}
        
        \label{tbl_agreement_first}
    \end{table}
    \end{minipage}\hspace{.7cm}
    \begin{minipage}[t]{.44\textwidth}
        \begin{table}[H]
        \caption{Inter-annotator agreement of three annotators A1, A2 and A3 on the two tasks after re-training with the updated annotation guidelines}
        \centering
        \begin{tabular}{|c|c|c|c|c|c|c|c|}
            \hline
            & \multicolumn{3}{|c|}{Task 1} & \multicolumn{3}{|c|}{Task 2}\\
            \hline
            & A1 & A2 & A3 & A1 & A2 & A3 \\
            \hline
            A1 & - & 0.93 & 0.59 & - & 0.87 & 0.57\\
            \hline
            A2 & - & - & 0.65 & - & - & 0.60 \\
            \hline
            A3 & - & - & - & - & - & - \\
            \hline
            \textit{Average $\kappa$} & \multicolumn{3}{|c|}{0.72} & \multicolumn{3}{|c|}{0.68} \\
            \hline
        \end{tabular}
        
        \label{tbl_agreement_second}
    \end{table}
    \end{minipage}
\end{figure}

According to Table \ref{tbl_agreement_first}, the average inter-annotator agreements of two tasks are lower than 0.5, which does not satisfy the minimum agreement level, according to \cite{di-eugenio-2000-usage}. Therefore, we update the current annotation guidelines with more explanations and examples and re-train the annotators to boost the quality of the annotation guidelines. As illustrated in Table \ref{tbl_agreement_second}, the inter-annotator agreement is improved on both tasks. The final inter-annotator agreement of our dataset is 0.72 and 0.68, which is the substantial level according to \cite{landis1977measurement}.

\begin{table}[H]
    \centering
    \caption{Confusion matrix between three pairs of annotators when annotating the samples. We calculate the values by taking the average of three pairs}
    \begin{tabular}{|c|c|c|c|c|}
        \hline
        & \textbf{NO SPAM} & \textbf{SPAM-1} & \textbf{SPAM-2} & \textbf{SPAM-3} \\
        \hline
        NO SPAM & \textbf{170.33} & 2.00 & 4.33 & 1.33 \\
        \hline
        SPAM-1 & 7.33 & \textbf{8.67} & 1.33 & 3.67 \\
        \hline 
        SPAM-2 & 12.00 & 1.33 & \textbf{12.00} & 0.00 \\
        \hline
        SPAM-3 & 11.00 & 1.00 & 5.00 & \textbf{58.67} \\
        \hline
    \end{tabular}

    \label{tbl_inter_conf_matrix}
\end{table}

In addition, Table \ref{tbl_inter_conf_matrix} describes the number of annotated comments by three different annotators. It can be seen that the number of disagreement data fell into the case of determining whether a comment belongs to a specific spam type and comment is not a spam review. Therefore, we attach the original links of products to the reviews for annotators to reference. This way, annotators can identify the reviews' context, then give the accrue labels. However, as shown in Table \ref{tbl_example_disagreement}, two annotators are disparity when annotating Reviews \#1, because this review contains the user's opinion about not only the product but also the brands of the providers, which is categorized as SPAM-2 according to the annotation guidelines. Besides, Comments \#2 and Comments \#3 give the opinion about the product's quality and design but do not mention the product carefully, which confuses the annotators for deciding these comments as non-review (SPAM-3).  

In general, the challenge of annotating for this task is identifying whether or not the reviews are spam. Despite detailed annotation guidelines and the information about the products relevant to the reviews, annotators are still misunderstood because consumers' opinions are diverse, and the stylistics of users' reviews are unclear. Therefore, to guarantee the objectives, we let three annotators annotate the entire dataset, then take the final label by major voting.

\begin{table}[H]
    \centering
    \caption{Several sample reviews that contain disagreement labels between annotators}
    \resizebox{1\textwidth}{!}{
    \begin{tabular}{c|p{8cm}|c|c}
        \textbf{No.} & \textbf{Reviews} & \textbf{Annotator 1} & \textbf{Annotator 2} \\
        \hline
        1 & Tiki giao hàng cực nhanh dù trong mùa dịch. Nivea xài tốt xưa giờ rồi. Tuy có hơi rít nhẹ, xài một thời gian da mềm hơn, sáng hơn. (\textit{\textbf{English}: Tiki delivered very quickly even during the epidemic season. Nivea used it very well from the past to now. Although there is a slight hiss, after using it for a while, the skin is softer and brighter.}) & SPAM & NO SPAM\\
        \hline
        2 & Hàng đúng loại, giá rẻ (\textit{\textbf{English}: Right product, cheap price}) & NO SPAM & SPAM \\ 
        \hline
        3 & Đẹp nha :)) sang xịn mịn (\textit{\textbf{English}: Beautiful :)) luxurious and smooth }) & NO SPAM & SPAM \\
        \hline
    \end{tabular}
    }
    
    \label{tbl_example_disagreement}
\end{table}

\subsection{Dataset Overview}

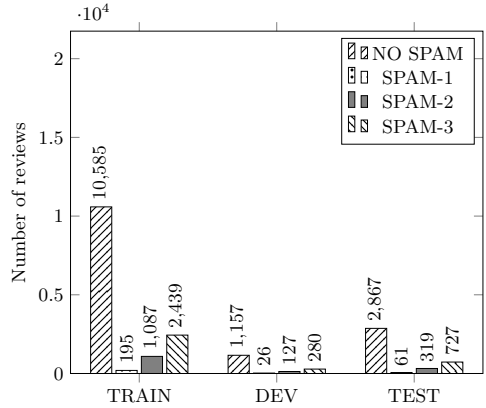
\begin{figure}[H]
    \centering
    \begin{tikzpicture}[scale=0.8]
            \begin{axis}[
                ybar,
                enlarge y limits={0.45,upper},
                enlarge x limits=0.25,
                symbolic x coords={TRAIN, DEV, TEST},
                xtick=data,
                ymin = 0, ymax = 15000,
                nodes near coords,
                every node near coord/.append style={rotate=90, anchor=west},
        	    ylabel near ticks,
        	    ylabel={Number of reviews},
        	    x tick label
        	   style={font=\footnotesize}
            ]
            \addplot[draw=black, pattern=north east lines] coordinates {
                (TRAIN,  10585) 
                (DEV, 1157) 
                (TEST, 2867)
            };
            \addplot[draw=black, pattern=dots] coordinates {
                (TRAIN, 195) 
                (DEV, 26) 
                (TEST, 61)
            };
            \addplot[draw=black, fill=gray] coordinates {
                (TRAIN, 1087) 
                (DEV, 127) 
                (TEST, 319)
            };
            \addplot[draw=black, pattern=north west lines] coordinates {
                (TRAIN, 2439) 
                (DEV, 280) 
                (TEST, 727)
            };
            \legend{NO SPAM, SPAM-1, SPAM-2, SPAM-3}
            \end{axis}
    \end{tikzpicture}
    \caption{The distributions of three labels on the train, development, and test sets}
    \label{fig_data_distribution}
\end{figure}

After annotating the dataset, we have nearly 19,000 reviews from users, in which each review is categorized as spam or not spam. If the reviews are spam, they consist of the types of spam. Then, we divided the dataset into train, development, and test sets with proportions 7-1-2. The overall information about the dataset is illustrated in Table \ref{tbl_overview_information}.

In addition, Figure \ref{fig_data_distribution} shows the distribution of reviews by each label on the train, development, and test sets. The reviews which are annotated as not spam account for the highest proportion. For the spam types, the SPAM-3 reviews are more than two remaining types. The data distribution on the training, development, and test set is similar.

\begin{figure}[H]
    \centering
    \begin{minipage}[t]{.48\textwidth}
        \begin{table}[H]
            \centering
            \caption{Overview about the dataset. The vocabulary size is computed on syllable level}
        \begin{tabular}{c|c|c|c}
            \hline
                & Train & Development & Test \\
                \hline
                Num. reviews & 14,306 & 1,590 & 3,974 \\
                \hline
                Vocabulary & 19,677 & 5,046 & 9,040 \\
                \hline
        \end{tabular}
        
        \label{tbl_overview_information}
        \end{table}
    \end{minipage}\hspace{.6cm}
    \begin{minipage}[t]{.38\textwidth}
        
        \begin{table}[H]
            \centering
            \caption{The distribution of spam and non-spam reviews based on the rating stars of users}
            \begin{tabular}{c|c|c|c|c|c}
            Stars & 1 & 2 & 3 & 4 & 5 \\
            \hline
            NO-SPAM & 213 & 93 & 193 & 292 & 9,793 \\
            \hline
            SPAM & 3 & 1 & 1 & 0 & 190 \\
            \hline
        \end{tabular}
        
        \label{tab_distribution_star}
        \end{table}
    \end{minipage}
\end{figure}

Besides, according to Table \ref{tab_distribution_star}, most spam reviews have 5 stars rated by users. For no spam reviews, although the distribution is more uniform from 1 to 4, most reviews are rated as 5 stars. Hence, the rating star for the product is not reliable information for expressing the opinions of users about the quality of the product.  
\section{Methodologies}
\label{methodologies}
\subsection{Task Definition}
The problem of spam review detection is categorized as the text classification task. This problem comprises two tasks: Task 1 is the binary classification task for classifying whether a review is spam or not spam, and Task 2 is the multi-class classification task for identifying the type of spam, which are one of three types as mentioned in Section \ref{dataset}.

\subsection{Word Embedding}
Word embedding is a vector space used to represent text data that can describe the relationship, semantic similarity, the context of data. In natural language processing, word representation plays a vital role in many downstream tasks, such as classification tasks. 
On the task of text classification, the empirical results from \cite{huynh-etal-2020-simple} showed that the fastText pre-trained embedding provided by \cite{grave-etal-2018-learning} obtained robust results when integrating with various deep neural networks on social media texts. Therefore, we choose the fastText word embedding \footnote{\url{https://fasttext.cc/docs/en/crawl-vectors.html}} for our empirical results. 

\subsection{Deep Neural Network Models}

\textbf{Text-CNN }\cite{kim-2014-convolutional}: Convolutional neural network (CNN) is a model combined with many different layers. CNN is often applied in computer vision to extract features of images for image classification and has achieved high performance than traditional approaches. In addition, CNN is also applied in natural language processing problems, typically text classification with the Text-CNN model. This model is based on convolutional architecture to extract valuable features from natural texts.

\textbf{LSTM }\cite{hochreiter1997long}: Long Short Term Memory (LSTM) is an improved model from Recurrent Neural Network (RNN). LSTM helps the model remember the previous information for a long time, which is a restriction faced by the RNN model. LSTM comprises three gates: input gate, output gate, and forget gate. The input gate selects information to add to the context, the output gate decides whether the input is necessary for the present, and the forget gate is used to remove information from the context when it is no longer needed. This model helps the model classify the text better because it can capture contextual information in the entire text.

\textbf{GRU }\cite{cho-etal-2014-properties}: Gated Recurrent Unit (GRU) is a variant of the LSTM model. This model has lower complexity than the LSTM. While the LSTM has three gates, the GRU has only two gates: the update and reset gates. The update gate determines if any past information is retained and used in the future, and the reset gate decides that past information should be kept and any information forgotten. The advantage of GRU is using fewer parameters during training and therefore uses less memory, and training time is faster than LSTM.

\textbf{Transformers} is an architecture that has been proposed in recent years and is currently in widespread use. The appearance of BERT \cite{devlin-etal-2019-bert} helps many downstream tasks in NLP attain high-performance results while training on a small dataset. BERT and its variances become the baseline approaches in many NLP tasks, which is called BERTology \cite{rogers-etal-2020-primer}.

In the Vietnamese language, there are two kinds of BERTology approaches: multilingual and monolingual models \cite{van2021investigating,huy-monolingual}. As a result, the monolingual obtained better results than the multilingual models for the text classification task \cite{van2021investigating}, and sequence-to-sequence task \cite{huy-monolingual}. Therefore, we applied two monolingual BERT models, including PhoBERT \cite{nguyen-tuan-nguyen-2020-phobert} and BERT4News \cite{nguyen2021nlpbk} for our problem of detecting spam reviews. 
\section{Empirical Results}
\label{results}

\subsection{Baseline Results}
We implement two tasks for the spam detection problem. First, we implement a binary classifier to classify reviews as spam or no spam. Second, we construct a classification model to determine spam types for the review. We adapt the Text CNN, LSTM, GRU models, and transformers with PhoBERT and BERT4News for our tasks. Finally, we will use the Accuracy and macro-averaged F1-score metrics to evaluate the performance of baseline models.
\newline

\begin{table}[H]
    \centering
    \caption{The empirical results of classification models on the dataset}
    \footnotesize
    \begin{tabular}{c c c | c c} 
        \hline
        \textbf{Model} & \multicolumn{2}{c|}{\textbf{Accuracy (\%)}} & \multicolumn{2}{c}{\textbf{F1-macro (\%)}}\\  
               & \textbf{Task 1} & \textbf{Task 2} & \textbf{Task 1} & \textbf{Task 2} \\
         \hline
         Text-CNN & 84.18 & 83.42 & 77.89 & 64.74 \\ 
         \hline
         LSTM & 82.97 & 83.35 & 77.24 & 66.58 \\
         \hline
         GRU & 83.50 & 82.84 & 77.67 & 66.51\\
         \hline
         \textbf{PhoBERT} & \textbf{90.01} & \textbf{88.93} & \textbf{86.89} & \textbf{72.17} \\
         \hline
         BERT4News & 86.39 & 86.20 & 86.16 & 62.62\\
         \hline
    \end{tabular}
    
    \label{tbl_result}
\end{table}

On classifying reviews as spam or not, the Text CNN model results on the test set with Accuracy and macro-averaged F1-score of 84.18\% and 77.89\%, better than the two models LSTM and GRU. Also, on this task, the PhoBERT achieved better results than BERT4News with Accuracy and macro-averaged F1-scores of 90.01\% and 86.89\%, respectively. For the spam type detection task, the PhoBERT model obtains 88.93\% for Accuracy and 72.17\% for macro-averaged F1-score, higher than BERT4News. The results of the evaluation between the models are described in Table \ref{tbl_result}.

Based on the results from the training and evaluation of the models, we can see that the classification of the evaluations using the Transformer models gives better performance than the deep neural network models. The PhoBERT model obtained the best performance on the two tasks. As for detecting spam types, the results of Accuracy and F1-score are more different than classifying reviews are spam or not because of the imbalance between the data labels. Besides, classifying the reviews as spam or not is less complicated than detecting the types of spam, so the results of this task are higher than the task of spam types detection.

\subsection{Error Analysis}
\begin{figure}[H]
    \centering
    \begin{minipage}[t]{.45\textwidth}
        \includegraphics[scale=0.4]{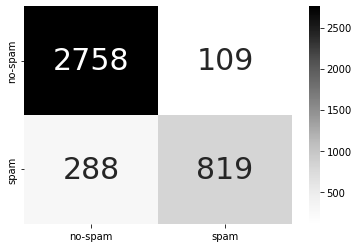}
        \small \centering
        Task 1: Detecting spam or no spam
    \end{minipage}\hspace{.6cm}
    \begin{minipage}[t]{.4\textwidth}
        \includegraphics[scale=0.4]{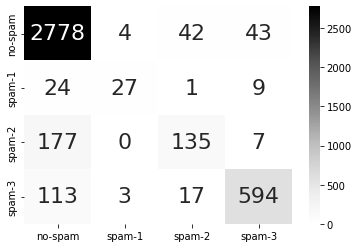}
        \small \centering
        Task 2: Identifying the type of spams
    \end{minipage}
    \caption{Confusion matrices of PhoBERT model on two tasks. The confusion matrices are created by using the sklearn library} 
    \label{fig_conf_matrix_errror}
\end{figure}

According to Fig \ref{fig_conf_matrix_errror}, the error prediction of SPAM and NO SPAM is not too much. In contrast, the error predictions are significantly different on the second task. Most of the SPAM-2 reviews are predicted as NO SPAM, and the number of wrong prediction (predicted as NO SPAM is 177) are higher than the accurate prediction (predicted as SPAM-2 is 135). The proportion of the wrong prediction of SPAM-1 reviews is also very high, in which most SPAM-1 reviews are predicted as no spam. However, this error prediction is not too much in the whole test set. The reviews with type SPAM-3 are the same as type SPAM-1. In general, most wrong predictions are caused by the doubt between NO SPAM label and other labels. Thus, the challenge of classification models on our dataset for this task is to determine whether the reviews are spam or not and to identify the type of spam reviews. 

\begin{table}[H]
    \centering
    \caption{Several wrong predictions of reviews with type SPAM-2 to NO SPAM label}
    \resizebox{1\textwidth}{!}{
    \begin{tabular}{c|p{8.5cm}|c|c}
        \textbf{No.} & \textbf{Reviews} & \textbf{True labels} & \textbf{Predicted labels} \\
        \hline
        1 & "Cảm ơn Tefal về chất lượng sản phẩm và cảm ơn Tiki về chất lượng dịch vụ. Chất lượng sản phẩm Tefal luôn vượt qua sự mong đợi của mình". (\textit{\textbf{English}: "Thank you Tefal for product quality and thank you Tiki for service quality. Tefal product quality always exceeds my expectations".}) & SPAM-2 & NO-SPAM \\
        \hline
        2 & hợp với giá tiền.giao hàng nhanh (\textit{\textbf{English}: suitable for the price.fast delivery}) & SPAM-2 & NO-SPAM \\
        \hline
        3 & Giao hàng nhanh, giá ổn (\textit{\textbf{English}: Fast delivery, good price}) & SPAM-2 & NO-SPAM \\
        \hline
        4 & 3 túi nước ariel 3,2 kg đã nhận.... chất lượng thì mình giặt đồ xong sẽ đánh giá sau. Thanhk Tiki và shop... (\textit{\textbf{English}: 3 bags of Ariel 3.2 kg laundry detergent received.... the quality will be evaluated after I wash the clothes. Thanks Tiki and shop...}) & SPAM-2 & NO-SPAM \\
        \hline
        5 & Pop It là đồ chơi xả stress theo lời con gái khoe. Shop giao hàng nhanh, trao đổi nhiệt tình (\textit{\textbf{English}: Pop It is a stress reliever according to what the girl said. Shop delivered products quickly, enthusiastic exchange}) & SPAM-2 & NO-SPAM \\
        \hline
    \end{tabular}
    }
    
    \label{tbl_example_error_analysis}
\end{table}

To study the wrong prediction in SPAM-2 labels, we take several random examples with predicted labels by the highest classification model and compare them with the real label. Those examples are described in Table \ref{tbl_example_error_analysis}. According to Table \ref{tbl_example_error_analysis}, there are two main reasons for the wrong predictions. The first reason is the identification of the praise for the brand of the product and the brand of the retailer or provider, and the reviews about the quality of products. For example, reviews No \#1, No \#3, and No \#4 mention the opinion about Tefal, Ariel (the two famous consumer goods brands in Vietnam), and brand Tiki (the online shopping service provider and retailer). However, those reviews do not focus on product quality, only express the thank to the retailer and the brands. The classification model cannot discriminate between the praise of the brands and the opinion of customers about the quality of products. The second reason is the short reviews of users, which are not giving any information about the products or services provider, such as reviews No \#2 and No \#3. 

In general, the cons of current classification models on the dataset are the perplexity of the spam and no-spam reviews, in which the reviews about products must directly focus on the characteristic of the product, the quality of the products, and their services. To overcome this problem, the model should integrate extra information about the product, such as the information page about the product and the previous reviews of users, to enhance the classification ability.

\section{Conclusion}
\label{conclusion}
This paper provided the ViSpamReviews - a dataset for spam reviews detection on Vietnamese online shopping websites with more than 19,000 reviews annotated by humans. The dataset follows the strict annotation process, and annotators are provided with detailed annotation guidelines for labeling the dataset. The final inter-annotator agreements are $\kappa=0.72$ for the task of determining whether a review is spam or not and $\kappa=0.68$ for the task of detecting the types of spam reviews. Besides, we also applied robust classification models to the dataset, and the PhoBERT model obtained the highest result with 86.89\% by F1-score for the spam classification task and 72.17\% by F1-score for the spam types detection task. From the error analysis, we found that it is necessary to integrate extra metadata about the product as well as the previous reviews to boost up the classification models.

Our next study is to extend the dataset for detecting spans of spam in the reviews and identify the opinion of users on the specific characteristic of products and their relevant services. Finally, based on the current results, the dataset can be used for developing an application to help shop owners filter spam reviews from users. 

\subsubsection*{Acknowledgements}
We would like to thank the annotators for their contribution to this work. This research was supported by The VNUHCM-University of Information Technology's Scientific Research Support Fund.

\bibliographystyle{splncs04}
\bibliography{references}

\end{document}